\documentclass[letterpaper, 10 pt, conference]{ieeeconf}
\usepackage[english]{babel}
\IEEEoverridecommandlockouts                              

\usepackage[letterpaper,top=2cm,bottom=2cm,left=3cm,right=3cm,marginparwidth=1.75cm]{geometry}

\usepackage{amsmath}
\usepackage{graphicx}
\usepackage{verbatim}
\usepackage[colorlinks=true, allcolors=blue]{hyperref}

\newcommand{\trav}{\tau}
\newcommand{\vect}[1]{\mathbf{#1}} 

\title{Complex Terrain Navigation via Model Error Prediction}

\author{Adam Polevoy$^{1*}$, Craig Knuth$^{1*}$, Katie M. Popek$^{1}$, Kapil D. Katyal$^{1}$
\thanks{$^{1}$Johns Hopkins University Applied Physics Laboratory, Laurel, MD, USA.
        {\tt\small Adam.Polevoy@jhuapl.edu, Craig.Knuth@jhuapl.edu}}%
\thanks{$^{*}$ These authors contributed equally.}
}

\begin{document}
\maketitle

\begin{abstract}
Robot navigation traditionally relies on building an explicit map that is used to plan collision-free trajectories to a desired target. In deformable, complex terrain, using geometric-based approaches can fail to find a path due to mischaracterizing deformable objects as rigid and impassable. Instead, we learn to predict an estimate of traversability of terrain regions and to prefer regions that are easier to navigate (e.g., short grass over small shrubs). Rather than predicting collisions, we instead regress on realized error compared to a canonical dynamics model. We train with an on-policy approach, resulting in successful navigation policies using as little as 50  minutes of training data split across simulation and real world. Our learning-based navigation system is a sample efficient short-term planner that we demonstrate on a Clearpath Husky navigating through a variety of terrain including grassland and forest.
 \end{abstract}

\section{Introduction}

\begin{figure}[]
	\centering
    \includegraphics[width=.95\columnwidth]{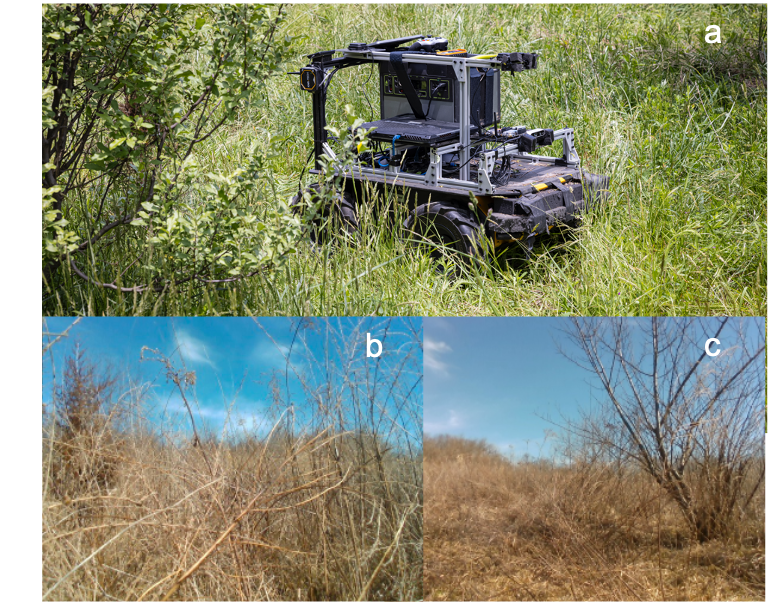}
    \caption{Our approach uses a deep neural network to estimate model error directly from onboard camera images. Primarily tested in a grassland (a), it is able to distinguish between  traversable tall grass (b) and impassable shrubs (c).}
	\label{fig:overview}
\end{figure}

Robot navigation in cluttered, complex environments remains a very challenging task. Traditionally, robots have built an explicit map of the environment with the goal of planning a collision-free path to a desired target position \cite{SLAMsurvey}. Although this approach has resulted in impressive results \cite{bresson2017SLAM}, it is difficult to build an explicit map in cluttered, unstructured environments due to occlusions and noisy sensor readings. To address this, we previously studied using deep neural networks to predict occupancy outside of a sensor's field of view \cite{katyal2019uncertainty}, \cite{katyal2021}, but the computational load of building an explicit map limited the robot's speed. Explicit mapping is further complicated in environments with deformable objects such as vegetation. Geometric-based approaches may not accurately reflect the environment, mislabeling tall grass as an impassable obstacle (see Fig.~\ref{fig:overview}). Also, geometric-based maps are usually unable to characterize terrain that may be difficult to navigate (i.e., ice, mud). This makes navigation through unstructured, outdoor environments extremely difficult in practice. 

In this work, we seek to plan collision-free paths directly from sensor data, bypassing the need to build an explicit map of the environment, even in complex, outdoor spaces. We utilize machine learning techniques to predict model error associated with an RGB image of the environment and a set of actions (i.e., trajectory). We frame this as a regression problem which estimates predicted error between the realized odometry readings and the predicted trajectory rollouts using a dynamical model of the robot. This model error encompasses errors due to faulty odometry, dynamical model mismatch, and environmental disturbances. Thus, in addition to learning to avoid nontraversable regions, our model also learns to avoid control regions resulting in high model uncertainty. The predicted error is scaled and combined with other relevant objectives in order to navigate without collision while achieving the task. A weighted average of the highest scoring trajectories is used with a gradient-free sampling-based trajectory planner \cite{nagabandi2020deep} to demonstrate successful navigation through multiple off-road environments.

The primary contributions of this paper include:

\begin{itemize}
    \item A novel deep neural network architecture for predicting model error in complex terrains
    \item Self-supervised data collection and training pipeline with minimal human intervention
    \item Real-world hardware experiments in complex terrain
\end{itemize}

\section{Related Work}
Traditionally, robotic navigation has followed a sense, plan, act loop, estimating the local environment in an occupancy map, planning collision-free paths, then controlling the robot to follow the desired trajectory\cite{smith1986, ayache1988, chatila1985}. While approaches that explicitly map the environment have yielded impressive results; they do have limitations, especially in unstructured, cluttered environments. Deformable objects (e.g., vegetation) may be misclassified as obstacles, causing the map to not improve over time. 

\subsection{Traversability Analysis}
Recently, there has been significant research into using machine learning to aid or replace portions of the navigation pipeline to help overcome these limitations \cite{ML_survey, xiao2020motion}. Several have trained classifiers to learn what terrain is traversable \cite{martinez2020, ahtiainen2017}. An alternative is to use semantic segmentation \cite{jackel2006, maturana2018} to distinguish types of terrain and prevent misclassification. Wellhausen et. al. \cite{Wellhausen19} propose simultaneously learning traversability prediction and semantic segmentation by using data from force-torque sensors embedded in a legged robot's feet during teleoperated runs. This traversability information is then used to update an underlying cost map. Our approach requires no map, instead directly planning trajectories from sensor data. Similar methods exist, which fuse perception and control, and plan trajectories directly from sensor data (e.g., point clouds \cite{gao2019flying} or images \cite{bansal2020, florence2020}), but these have not previously considered terrain traversability. 

\subsection{Motion Prediction}
Guzzi et. al. \cite{Guzzi20} predict the probability that a robot will be able to navigate from one pose to another using a depth map of the environment, then chain the predictions together to form a trajectory. Our approach instead uses a known canonical dynamics model of the robot and samples feasible trajectories with corresponding estimates of model error. Rather than estimating the confidence of an existing controller, we assume a dynamics model is known and learn deviations from that model. Although this model may not be accurate in all terrains any error is incorporated into our prediction causing our system to prefer terrain in which the model performs better. 

\subsection{End-to-End Navigation Systems}
Several end-to-end learning systems for navigation exist, both in structured urban environments \cite{chen2015} and those focused on off-road terrain \cite{kahn2021badgr, pan2020, muller2006}, but generally these require significant training data.

This work is most closely related to the end-to-end navigation system called Berkeley Autonomous Driving Ground Robot (BADGR) first proposed in \cite{kahn2021badgr} with a few key differences. BADGR allows for robots to reason about navigation directly from image data using a learned neural network model. Specifically, it learns to classify events, focusing on collisions and bumpiness. It also regresses position from action sequences and image data. Instead of relying on a learned model for the complete navigation pipeline, we estimate our position using GPS and focus on estimating model error for a trajectory based on regressed model error. As we are not trying to implicitly learn a model of the robot's dynamics, our method requires much less training data and removes the need for human-engineered functions for event classification. Further, it enables us to estimate traversability along an entire trajectory with a single scalar value.

\begin{figure}[]
	\centering
    \includegraphics[width=1\columnwidth]{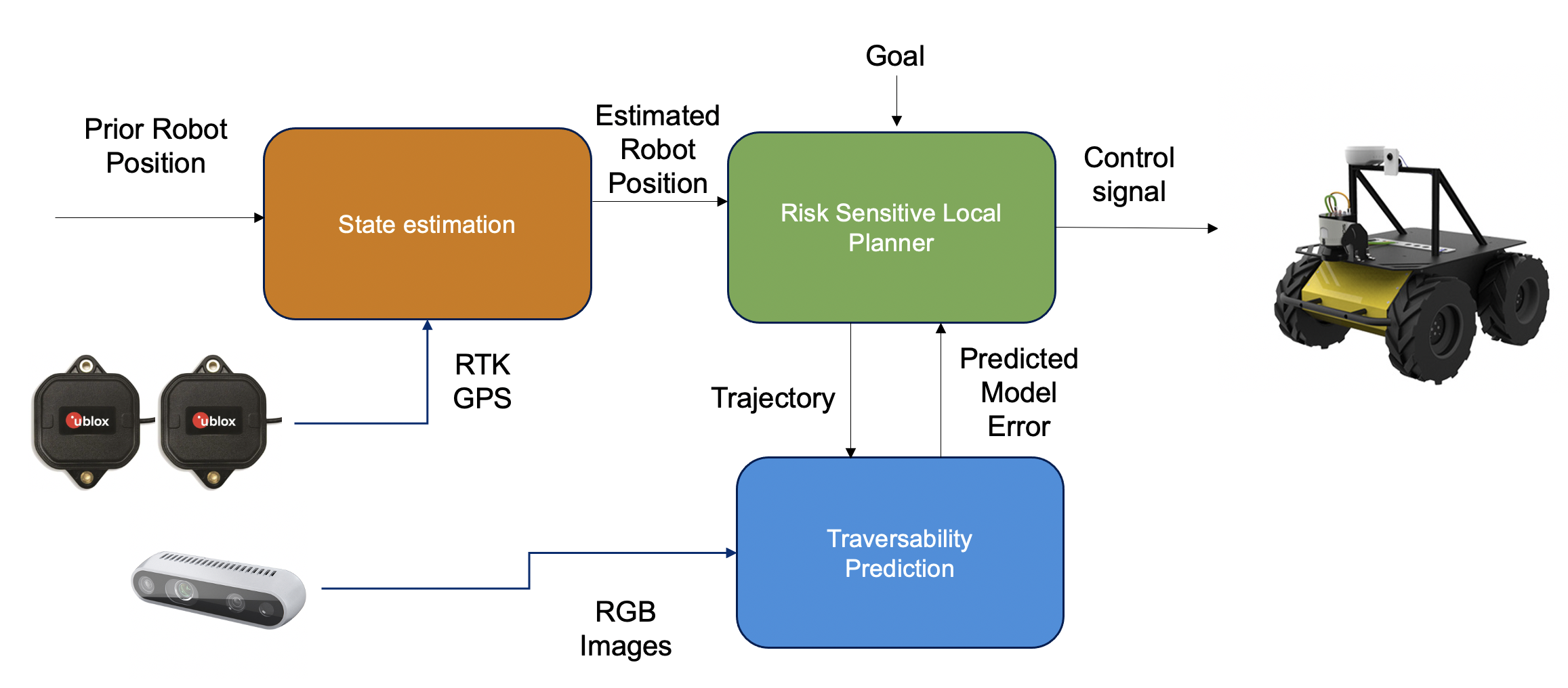}
    \caption{Overview of our approach. This paper focuses on the blue Model Error Prediction box which takes two RGB images and set of trajectories as input and outputs the corresponding estimated error. This prediction is paired with two RTK GPS units for state estimation and a sampling-based controller \cite{nagabandi2020deep}.}
	\label{fig:approach}
\end{figure}
\section{Approach}

\subsection{Problem Formulation}

Our objective is to regress model error using sensor data and an action trajectory. This error reflects the robot's ability to reliably execute a sequence of actions. We then aim to utilize this error, in tandem with task relevant objectives, to inform robot navigation. A high-level overview of our approach is shown in Figure~\ref{fig:approach}.

We choose a canonical discrete model $x_{t+1} = f(x_t, u_t)$ of the robot's dynamics with $x$ being the robot state and $u$ being the control. Suppose an action trajectory, $\vect{u} = \{u_0, \ldots, u_H\}$, is executed and results in some measured realized state trajectory $\vect{\hat{x}} = \{\hat{x}_0, \ldots, \hat{x}_{H+1}\}$ where $H$ is the horizon. Let $\vect{x}$, defined similarly, be the canonical state trajectory as determined by the canonical model starting from the same state as the measured realized trajectory (i.e. $x_0 = \hat{x}_0$, $x_{k+1} = f(x_{k}, u_{k})$). We define the model error as the maximum state error between the realized and canonical trajectories:
\begin{equation}\label{eq:model_err}
    \trav(\vect{\hat{x}}, \vect{u}) = \underset{0 \leq k \leq H+1}{\max} \|x_k - \hat{x}_k\|
\end{equation}
where $\|\cdot\|$ is the standard $L_2$ norm.

To control the robot, we maximize a reward function that is the sum of an exponential transformation of the model error and other relevant task objectives.
\begin{figure*}[t]
	\centering
    \includegraphics[width=1.90\columnwidth]{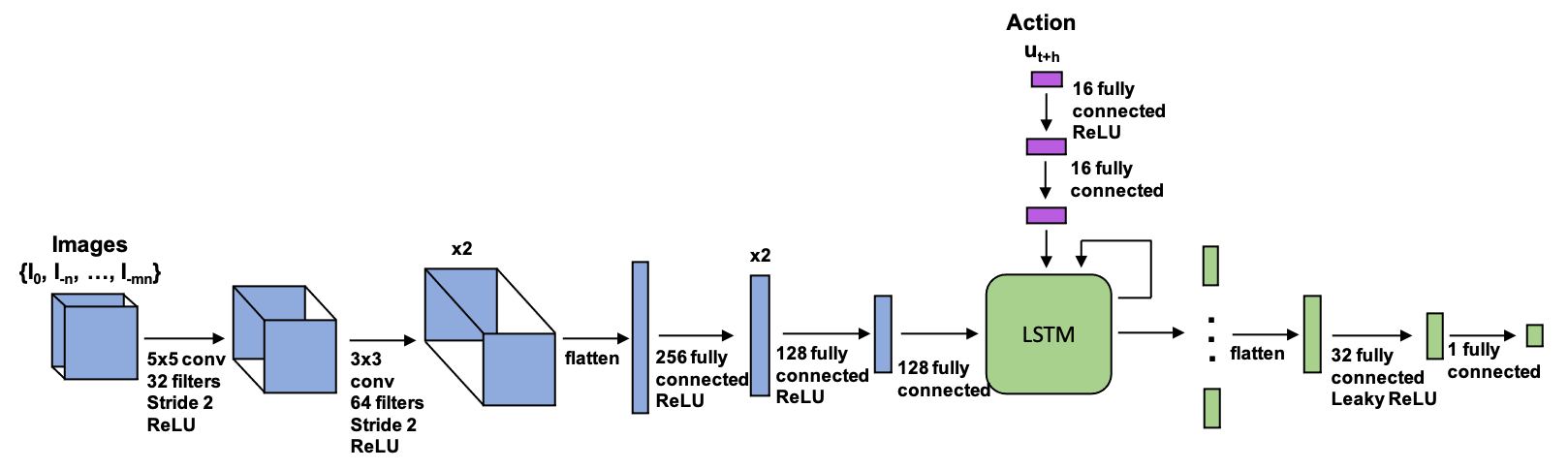}
    \caption{The neural network architecture used for our model error prediction. It features two subnetworks, the first is a CNN that takes a history of RGB images as input. The second subnetwork is a fully connected network that take actions as input. The latent feature vectors from these two networks are used as input into an LSTM. Our output is a single error estimate. }
	\label{fig:network}
\end{figure*}
However, the model error cannot be computed online as it requires future measurements of the realized trajectory. Instead, we regress the model error using an image history and the action trajectory. Specifically, let $\vect{I} = \{I_0, I_{0-n}, \ldots, I_{0-mn}\}$ be a history of $m+1$ images spaced at $n$ timesteps apart. We approximate $\trav(\vect{\hat{x}},\vect{u})$ with a neural network $\tilde{\trav}_\theta(\vect{I},\vect{u})$ that is parameterized with variables $\theta$. 

This formulation of the regression target encodes multiple sources of error, including environmental disturbances, dynamical model mismatch, and odometry error, into a single metric per trajectory. Further, it does so without learning the dynamical model itself, which would have more significant training data requirements.

\subsection{Model Error Prediction}

Our error prediction model is a two prong neural network (Figure \ref{fig:network}), similar to that of \cite{kahn2021badgr}, that regresses on the model error (Eq. \eqref{eq:model_err}). One subnetwork is a Convolutional Neural Network (CNN) that takes a history of RGB images as input. The other subnetwork is a fully connected network that takes an action sequence as input. Outputs from these subnetworks are processed through a Long short-term memory (LSTM) network and a fully connected network.

Our network has some key differences from that used in \cite{kahn2021badgr}. First, to capture temporal features in the image embedding, our image subnetwork takes in a history of images, rather than only the current image. Second, we flatten the output of the LSTM into a single feature vector, representing the entire trajectory, rather than knot points along the trajectory. This allows our network to regress a single error estimate per action sequence, whereas the network in \cite{kahn2021badgr}, classifies and regresses multiple metrics for each action of each action sequence. Additionally, we observed improved performance by switching from Rectified Linear Unit (ReLU) to leaky ReLU activation functions on the final layers of the network. 

We use a mean squared error loss between the predicted trajectory and observed odometry, where $\theta$ represents the network parameters.
\begin{equation}
    L(D; \,\theta) = \frac{1}{N} \sum_{i=0}^{N} \| \trav(\vect{\hat{x}}^i, \vect{u}^i) - \tilde{\trav}_\theta(\vect{I}^i, \vect{u}^i)\|^2 \nonumber
\end{equation}



\subsubsection{Dataset Collection}

\begin{figure}[]
	\centering
    \includegraphics[width=0.9\columnwidth]{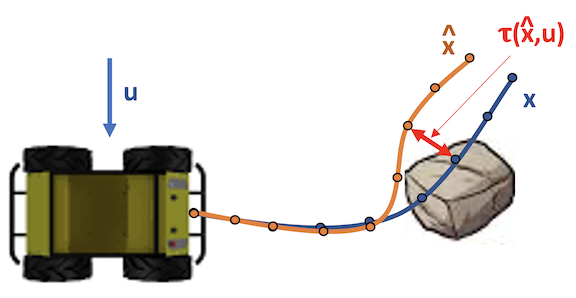}
    \caption{Our training dataset is created in a self-supervised manner with limited human interaction. The predicted trajectory using the canonical model, $\textbf{x}$, is compared with the realized trajectory, $\hat{\textbf{x}}$, to get a model error $\tau$ (Eq.~\ref{eq:model_err}).}
	\label{fig:dataset_gen}
\end{figure}

The input and target data for model error regression is generated in a self-supervised manner. This is necessary to enable efficient collection of sufficiently large datasets. During data collection, as the robot navigates representative terrains, odometry, action commands, and images are recorded. Once data collection has finished, the odometry and action commands are sub-sampled and time aligned to match the image data via linear interpolation. The generation of model errors from the dataset for use as regression targets is shown in Figure \ref{fig:dataset_gen}.




While random control sequences provide an easy method to drive the robot for data collection, the control action distribution will not match the controller's during execution. Instead, the model will frequently be trained on action sequences that are never executed in practice and without significant training data, informative control action sequences will be missed. Teleoperation is an alternative approach, but human interaction is costly and time consuming. Instead, we collect  data in an on-policy fashion, with the controller utilizing the model error prediction, ensuring performance is tailored to the data distribution seen in practice. Using the controller in this manner requires a learned perception model and data must be collected prior to training. This results in an alternating process between training the model and collecting data until a satisfactory performance threshold is achieved.

To achieve this, we propose an iterative data collection and network training pipeline. An initial dataset can be collected through either human teleoperation or random control sequences, and an initial perception model can be trained using this data. Then this model is used with the controller to collect more data by generating random goal positions and allowing the controller to use the learned perception model to navigate to the goal. Once the system reaches the goal, or a time limit is reached, a new goal position is generated. The newly collected data is added to the dataset, and the model is retrained starting from the current network weights. Data collection and retraining can then be iterated upon until performance is satisfactory. Human interaction is only needed during the training process if the robot encounters a nontraversable obstacle and needs to be reset.

\subsection{Controller}

Our controller is a gradient-free model predictive path integral (MPPI) controller developed for use with deep neural networks \cite{nagabandi2020deep}. Given a reward function, it will iteratively sample perturbations from a nominal trajectory, compute a reward-weighted average of those samples as an estimate of the optimal trajectory, execute the first action of the trajectory, then repeat.

Our reward function is comprised of two parts: a goal reward and a traversability reward. The goal reward was defined as the weighted sum of distance traveled in the direction of the goal $x_g$ according to the canonical model:
\begin{equation}
R_g(x_0, \vect{u}) = \sum_{i=0}^{H} \alpha (x_{i+1} - x_i) \cdot \frac{x_g - x_i}{\|x_g - x_i\|}
\end{equation}
where $\alpha$ is a scaling factor. For ease of scaling, this reward is normalized by dividing by the maximum possible distance traveled towards goal. For Dubins car dynamics, this maximum distance results from the robot driving directly towards the goal at a maximum velocity $v_\text{max}$, resulting in the normalization factor $\eta = 1 / (v_{\text{max}} H dt)$, where $H$ is the command horizon. After normalization, it is clipped to the range $[0,1]$, yielding:
\begin{equation}
\begin{aligned}
    R^n_g(x_0, \vect{u}) &= 
\begin{cases} 
      0, & \eta R_g(x_0, \vect{u}) \leq 0 \\
      1, & \eta R_g(x_0, \vect{u}) \geq 1 \\
      \eta R_g(x_0, \vect{u}), & \text{else}
\end{cases}
\end{aligned}
\end{equation}

The traversability reward function is calculated using the predicted model error. First the network output is normalized similarly to $R_g^s$ (but not clipped). Then, it is fed through an exponential map only if larger than some bias $\beta$.
\begin{equation}
\begin{aligned}
    R_\trav(\vect{I}, \vect{u}) =
    \begin{cases}
          0, \quad \eta \tilde{\trav}_\theta(\vect{I}, \vect{u}) \leq \beta \\
          -e^{\sigma(\eta \tilde{\trav}_\theta(\vect{I}, \vect{u}) - \beta)} - 1, \quad \text{else}
    \end{cases}
\end{aligned}
\end{equation}
where $\sigma$ is an exponential scaling factor.

In practice, we found that a linear mapping of the predicted model error was not sufficient to avoid dangerous situations. By exponentially mapping the error, the robot is able to quickly react to obstacles appearing in the scene and only rely on the goal reward when trajectories can be effectively executed. 

We empirically selected $\beta$ and $\sigma$. Due to factors like model mismatch and state estimation errors, we found there always existed some level of baseline error between the realized and canonical state trajectories. We selected the bias, $\beta$, to offset the reward so the controller only reacts to predictions that exceed this baseline, such as in the presence of obstacles. $\sigma$, along with $\alpha$ in the goal reward, can be freely chosen to balance the sensitivity between the goal reward and the traversability reward.

The final reward function utilized by the controller is the sum of the goal and traversability rewards:
\begin{align}
    R(x_0, \vect{I}, \vect{u}) &= R_g^n(x_0, \vect{u}) + R_\trav(\vect{I}, \vect{u})
\end{align}

\section{Experimental Evaluation}

\begin{figure}[]
	\centering
    \includegraphics[width=0.9\columnwidth]{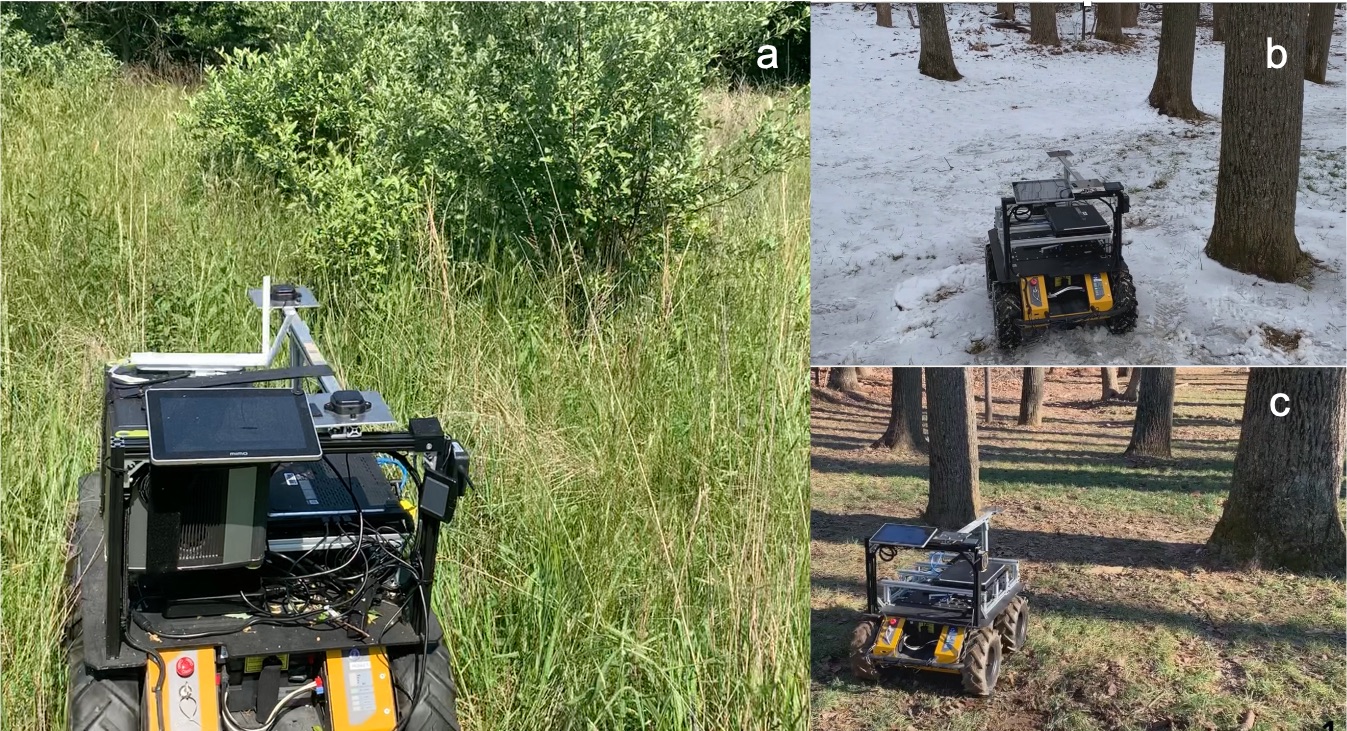}
    \caption{Grassland was our main test environment (a). Our method also successfully traversed through trees (b,c).}
	\label{fig:environment}
\end{figure}

Our approach was trained and primarily tested in a grassland environment, shown in Figure \ref{fig:environment}a. This environment was particularly applicable to our approach given the combination of deformable (e.g., tall grass) and nondeformable  vegetation (e.g., shrubs). In addition to grasslands, we also tested navigated through trees (Fig.~\ref{fig:environment} b,c).

\subsection{Hardware}

We evaluate our method on a Clearpath Husky\textsuperscript{\texttrademark}.
The controller and network computation is performed on an ASUS SCAR Strix G532LWS XS96 which has Core i9 CPU and GF RTX 2070 SUPER GPU. 
We utilize an Intel Realsense D435i 
for RGB images (ignoring depth and IMU readings), and two ublox Real-time Kinematic GPS units for odometry. 

\subsection{Canonical Model and Parameter Selection}

We selected our canonical model to be a discretized version of Dubins car dynamics. The state is comprised of $\chi$, $y$, and a heading $\phi$, while the input is forward and angular velocities, $v$ and $\omega$. The resulting dynamics are:

\begin{equation}
\begin{aligned}
    \chi_{t+1} &= \chi_t + dt \, v \cos \phi_t \\
    y_{t+1} &= y_t + dt \, v \sin \phi_t \\
    \phi_{t+1} &= \phi_t + dt \, \omega
\end{aligned}
\end{equation}

We selected $dt = 0.1$ and $H = 20$ corresponding to a two second time horizon. We used two images in the image history, the current image and the image from time $t-1$ (s). We choose $\beta = 0.1$, $\sigma = 10$, and $\alpha = 1$. For the controller, we sampled $128$ trajectories with a sampling covariance of $[0.25 \, 0; 0 \, 0.25]$, smoothing factor of $0.5$, and reward weighting factor of $50$ (corresponding to $\Sigma$, $\beta$, and $\gamma$ respectively in \cite{nagabandi2020deep}). Additionally, $v_\text{max} = 0.8$ m$/$s for the Husky.

\subsection{Training}

\begin{figure}[]
	\centering
    \includegraphics[width=0.9\columnwidth]{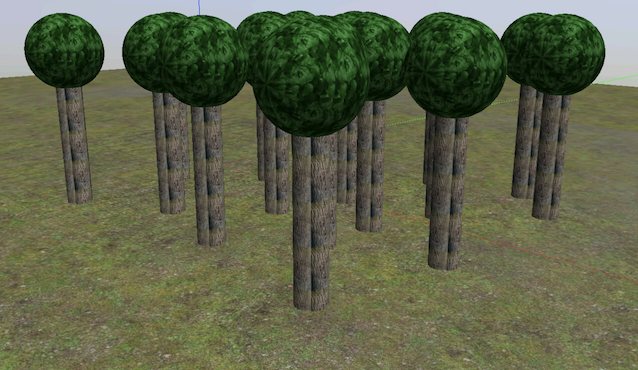}
    \caption{Gazebo simulation environment with tree obstacles.}
	\label{fig:simulation}
\end{figure}

The weights of our model error estimation network were initialized from data collected in simulation. The simulation was developed in Gazebo \cite{gazebo04} and consists of several cylindrical tree obstacles, as shown in Figure \ref{fig:simulation}. The Husky was represented using a box and a Dubins car model. The network was trained on a total of 20 minutes of data collected across 5 iterations, in which it learned to avoid the trees while navigating to the goal.

\begin{figure}[]
	\centering
    \includegraphics[width=0.9\columnwidth]{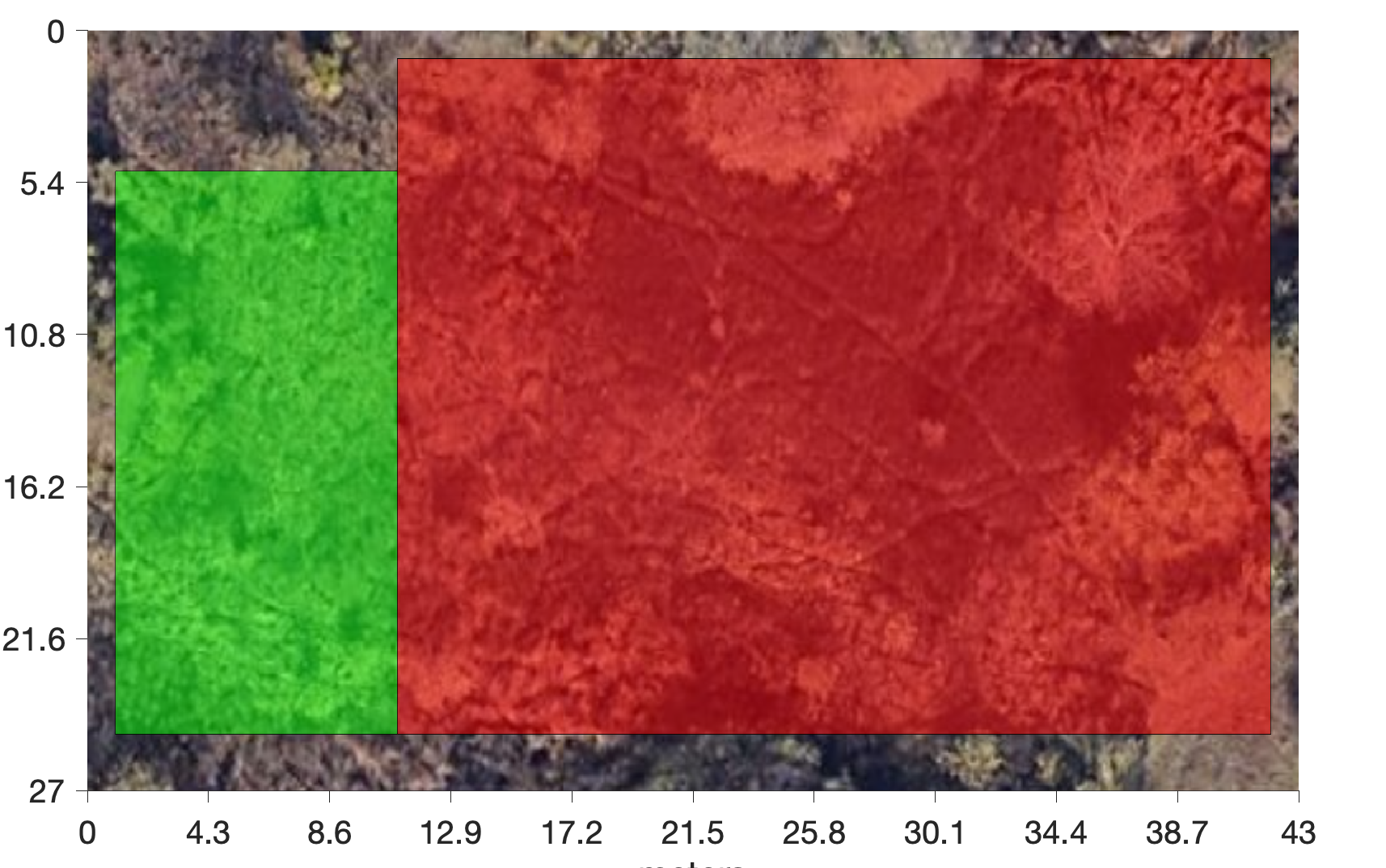}
    \caption{The training area is depicted in green, and the testing area is depicted in red.}
	\label{fig:train_test}
\end{figure}

The grasslands environment was divided into training and testing sections, as shown by the green and red rectangular overlays in Figure \ref{fig:train_test}. We collected an initial 15 minute dataset by teleoperating the robot in the training area to capture representative scenarios such as colliding into bushes, driving around obstacles, and driving through tall grass. We then trained the network, which was initialized with the simulation dataset. The network was trained with a 0.9/0.1 train/validation split for a maximum of 20 epochs. The training was terminated if the validation loss plateaued (i.e. 3 consecutive epochs with a change of less then 1e-4).

After initial training, we recollected data for three minutes with autonomous control rather than manual operation, repeating this five times for a total of 30 minutes of real world data.

\subsection{Experimental Setup}


In the grassland environment, after 30 minutes of real world data collection, we tested the system by navigating to five waypoints, spread across approximately 70\,m. Four of the waypoints were in the testing area not seen during training, which contained different types of vegetation, including deformable tall grass and stronger immovable bushes.  The waypoint positions relative to the initial position of the robot were [(-3.6, 15.4), (10.3, 14.2), (25.7, 14.0), (23.7, 3.89), (34.7, -4.9)]. An experiment was considered successful if the system reached all waypoints without getting stuck on obstacles, as shown in the attached supplementary video.

\subsection{Experimental Results}
\begin{figure*}[th]
	\centering
    \includegraphics[width=1.9\columnwidth]{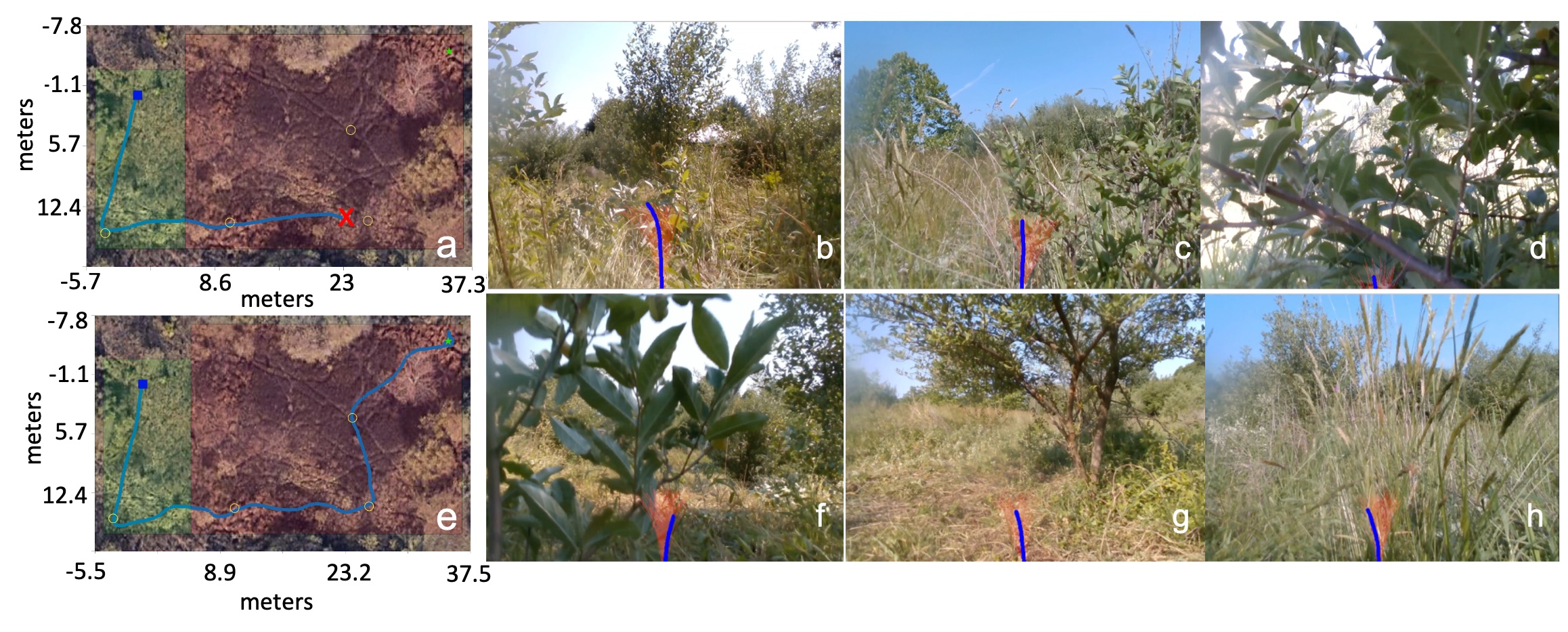}
    \vspace{-5mm}
    \caption{Left column shows an overhead view of the trajectory for Grassland Trial 1 (top) and Grassland Trial 2 (bottom). The training area is represented in green, and the testing area in red. The blue trajectory is the path traversed by the robot. The blue square is the start point, the yellow circles are intermediate waypoints, and the green star is the goal. Axes values are distance in meters. Right columns show sampled onboard images from the corresponding trajectory and depict the variation in traversed terrain. The corresponding sampled and chosen trajectories for a two second time horizon are overlaid in red and blue respectively.}
	\label{fig:results}
\end{figure*}
In addition to the the multi-waypoint demonstrations in grassland, we share the statistics of the system as it was trained in Table \ref{tab:training}. As mentioned previously, the on-policy training sessions occur after manual data collection and initial training. The training region is approximately 10m x 20m. As we randomly sampled goals during training, some variability exists in the number of goals achieved during each iteration. The number of collisions was recorded by a human during the execution of the test.
\begin{table}[ht]
    \centering
    \begin{tabular}{c || c | c | c | c | c}
        Training Iteration & 1 & 2 & 3 & 4 & 5 \\
        \hline
        Goals Reached & 5 & 6 & 5 & 4 & 8 \\
        Collisions & 1 & 2 & 3 & 2 & 0
    \end{tabular}
    \caption{Goals reached and collisions during training.}
    \label{tab:training}
\end{table}

During the first grassland trial, the system was successfully able to navigate through the training area, but failed to navigate around a bush halfway through the novel environment. The navigated trajectory,  and a sample of the corresponding onboard camera images are shown in the top row of Figure \ref{fig:results}(a-d). The failure, depicted with a red `x', was caused by the bush shown in Figure \ref{fig:results}d. 

For the second grassland trial, we retrained the network with the training dataset augmented by data collected in the first trial. With this small addition, the system was able to successfully navigate to all five waypoints, demonstrating our ability to continually improve performance with new experiences. The resulting trajectory and sampled onboard images are shown in the bottom row of Figure \ref{fig:results}(e-h). 

\subsection{Discussion}

This method enjoys some natural advantages in generalization to novel environments. Due to the formulation of the model error, the neural network regresses only positive values, therefore uncertain input generally yields larger positive predicted error. To test our method in novel situations, the network was trained in a tree environment (Fig.~\ref{fig:environment}c) and tested without any additional training after a snowfall (Fig.~\ref{fig:environment}d). While it was more cautious, it was able to successfully navigate to our goal as demonstrated in the supplementary video.

Our formulation allows us to encode multiple sources of model error (e.g. dynamical model mismatch, environmental disturbances, odometry error) without needing to learn the dynamical model itself. Additionally, learning the model error from on-policy data further reduces the data requirement. In the grassland environment, we required only 50 minutes of data (20 minutes simulated and 30 minutes real world), far less than comparable methods. Finally, it removes the need to explicitly identify adverse training events, (e.g., collisions), resulting in a more intuitive and salient regression target.

\section{Conclusion}

In this paper, we have presented a learned perception model capable of predicting model error for use in robot navigation. Specifically, we developed a neural network to estimate the maximum model error resulting from a commanded trajectory given sensor data.  We also showcase a methodology for collecting on-policy training data, with the controller in the loop, for efficient data collection with minimal human interaction. Further, we developed a reward function and integrated an MPPI controller to showcase real-time control with our learned perception model. We demonstrated successful navigation on a physical robot using the perception model trained on a small dataset of only one hour. Even with this limited training set, our system achieved robust navigation in complex terrains over long distances ($\approx$70\,m), see supplementary video.

Despite positive results, there is much room for future work. As shown in our results, while generalization to novel environments shows some promise, it is clear that further adaptation is needed. In this instance, approaches such as lifelong and continual learning could yield improved results, especially those that automatically trigger model retraining and quantification of confidence in model prediction.

Another key challenge is the requirement of "negative" examples, i.e., collisions where the model error is large. Ideally, a system would need few of these examples and would avoid situations where it is particularly costly, such as high-speed collisions. In this case, synthetic data creation, such as utilizing low-speed collisions to produce examples of high-speed collisions, may reduce the need for nsegative examples. In spite of these challenges, we believe learned policies have significant potential to improve navigation in complex, off-road terrains.

\bibliographystyle{IEEEtran}
\bibliography{main}

\end{document}